\documentclass[a4paper,11pt,twoside]{article}
\usepackage{fancyheadings}
\usepackage{graphicx}
\usepackage{url}
\usepackage{cite}

\def\id{??}   

\providecommand{\mictitle}[1]{\title{\Large\textbf{#1}}}
\providecommand{\firstaffiliationmark}{$^{\ast}$}

\providecommand{\micauthor}[3]{\index{{#2, #1}}#1 #2#3}
\providecommand{\micinstitution}[4]{\normalsize #1#2 \\ #3 \\ \texttt{#4} \\ \hfill \\ }
\providecommand{\institutions}[1]{\date{#1}}

\begin{document}
%
%

\mictitle{MOOPPS: An Optimization System for Multi Objective
Production Scheduling}

\author{
    \micauthor{Martin J.}{Geiger}{\firstaffiliationmark}
}

\institutions{
\micinstitution{\firstaffiliationmark}
    {Lehrstuhl f\"{u}r Industriebetriebslehre (510A), Universit\"{a}t Hohenheim}
    {Schlo\ss{} Hohenheim, Osthof-Nord, D-70593 Stuttgart, Germany}
    {mjgeiger@uni-hohenheim.de}
}

\maketitle
\thispagestyle{fancyplain}

\section{Introduction}
The resolution of multi objective optimization problems is
twofold: First, the set of Pareto-efficient alternatives $P$ with
respect to the defined objective functions has to be determined.
Second, an alternative $x^{*} \in P$ has to be chosen by the
decision maker. Obviously, practical problems require that both
aspects are addresses within a solution concept.

In the current paper, we present an optimization system solving
multi objective production scheduling problems (MOOPPS). The
identification of Pareto optimal alternatives or at least a close
approximation of them is possible by a set of implemented
metaheuristics. Necessary control parameters can easily be
adjusted by the decision maker as the whole software is fully menu
driven. This allows the comparison of different metaheuristic
algorithms for the considered problem instances. Results are
visualized by a graphical user interface showing the distribution
of solutions in outcome space as well as their corresponding Gantt
chart representation.

The identification of a most preferred solution from the set of
efficient solutions is supported by a module based on the
aspiration interactive method (AIM) \cite{lotfi:1992:article}. The
decision maker successively defines aspiration levels until a
single solution is chosen.

After successfully competing in the finals in Ronneby, Sweden, the
MOOPPS software has been awarded the \emph{European Academic
Software Award 2002} (\url{http://www.easa-award.net/},
\url{http://www.bth.se/llab/easa_2002.nsf}).

\section{\label{section:moscheduling}Multi objective production scheduling}
Production scheduling can be characterized as the assignment of
jobs $\mathcal{J} = \{ J_{j}, \ldots, J_{n}\}$, each of which
consists of a set of operations $J_{j} = \{ O_{j1}, \ldots,
O_{jo_{j}} \}$ to a set of machines $\mathcal{M}=\{M_{1}, \ldots,
M_{m}\}$. Processing of operations on the machines is done
involving a nonnegative processing time $p_{jk}$ for each
operation $O_{jk}$. A schedule $x$ defines starting $s_{jk}$ times
of the operations $O_{jk}$ on the machines. Based in this
assignment, completion times $C_{j}$ of the jobs $J_{j}$ are
derived.\\Typical side constraints that have to be taken into
consideration are precedence constraints among operations of jobs
and release dates $r_{j}$ of jobs $J_{j}$. Also, due dates $d_{j}$
may be present for each job $J_{j}$. An overview is given e.\,g.
in \cite{pinedo:2002:book}.

The schedule identified for a given problem should be of overall
maximum quality from the perspective of a so called \emph{decision
maker/planner/scheduler}. Often, multiple aspects or \lq{}points
of view\rq{} \cite{roy:1985:book} are of relevance that formally
can be expressed by a set of optimality criteria. For each
schedule $x$, a vector of objective function values $G(x) = \left(
g_{1}(x), \ldots, g_{k}(x) \right)$ determines its quality.
Important objective functions include the maximum completion time
or makespan $C_{max} = \max ( C_{j})$ of the jobs $J_{j}$, the sum
of the completion times $C_{sum} = \sum C_{j}$, the maximum
tardiness $T_{max} = \max ( T_{j} )$ with $T_{j} = \max ( C_{j} -
d_{j}, 0)$ and the number of tardy jobs $U=\sum U_{j}$ with
$U_{j}=1$ if $C_{j} > d_{j}$, 0 otherwise. Without loss of
generality, we assume in the further explanations that all
considered objective functions have to be minimized.

The goal of a multi objective optimization problem can be
formulated as to
\begin{equation}\label{MOP}
    ``\min" \,\, G(x)=\big{(}g_{1}(x),...,g_{k}(x)\big{)}
\end{equation}
$x\in\Omega$ as a solution of the problem and belongs to the set
of all feasible solutions $\Omega$. As often conflicting objective
functions $g_{k}(x)$ are considered, minimization does not lead to
a single optimal solution but is understood in the sense of
efficiency (or Pareto optimality)
\cite{vanveldhuizen:2000:article}.

\newtheorem{definition}{Definition}
\begin{definition}[Pareto dominance]\label{def1}
    An objective vector $G(x)$ is said to dominate $G(x')$,
    if $g_{i}(x)\leq g_{i}(x')\forall i\in \{ 1,...,k\}\wedge
    \exists i\in \{1,...,k\} \mid g_{i}(x)<g_{i}(x')$. We denote
    the domination of a vector $G(x)$ to the vector $G(x')$
    with $G(x)\prec G(x')$.
\end{definition}
\begin{definition}[Pareto optimality, Pareto set]\label{def2}
    A solution $x\in \Omega$ is said to be efficient or Pareto optimal,
    if $\neg\exists x'\in \Omega |x'\prec x$. The set set of all
    solutions fulfilling this property is called the Pareto set $P$.
\end{definition}
From the description of the multi objective optimization problem
in Expression (\ref{MOP}) we derive in combination with the
Definitions \ref{def1} and \ref{def2} the final goal to find all
$x\in P$. Finally, the decision maker is able to select a most
preferred solution $x^{*} \in P$.

\section{\label{section:dss}A decision support system for multi objective scheduling}
\subsection{System description}
\begin{figure}[ht!]
\centerline{\includegraphics{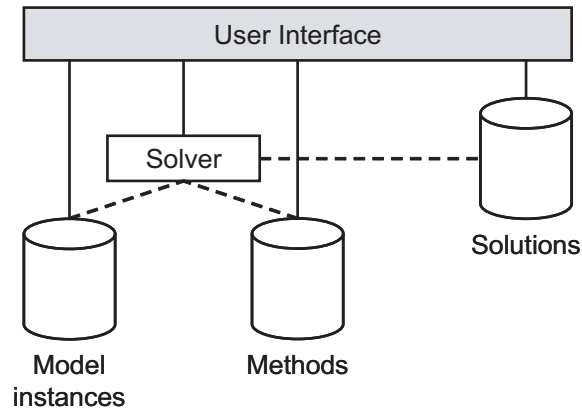}}
\caption{\label{fig:dss}Structure of the DSS
\cite{sprague:1982:book,pinedo:1997:article}.}
\end{figure}

For the resolution of multi objective production scheduling
problems, the integrated system MOOPPS has been implemented. As
illustrated in Figure \ref{fig:dss}, the system consists of
different components for the resolution of the problem.

A \emph{method database} contains a set of heuristics approaches
for solving multi objective scheduling problems:
\begin{enumerate}
    \item Priority rules \cite{haupt:1989:article}, based on the early work of Giffler and
    Thompson \cite{giffler:1960:article} for generating active
    schedules.
    \item Local search neighborhoods \cite{reeves:1999:article} within a multi-point
    hillclimber.
    \item Multi objective evolutionary algorithms
    \cite{bagchi:1999:book}, incorporating elitist strategies and
    a variety of crossover neighborhoods like e.\,g. uniform order
    based crossover, order based crossover, two point order
    crossover, and partially mapped crossover.
    \item The \lq{}MOSA\rq{} multi objective simulated annealing algorithm
    of Teghem et al. \cite{ulungu:1999:article}.
    \item A module based on the \lq{}AIM\rq{} aspiration interactive
    method \cite{lotfi:1992:article} for an interactive search in the obtained results.
\end{enumerate}

\begin{figure}[ht!]
\centerline{\includegraphics[width=14cm]{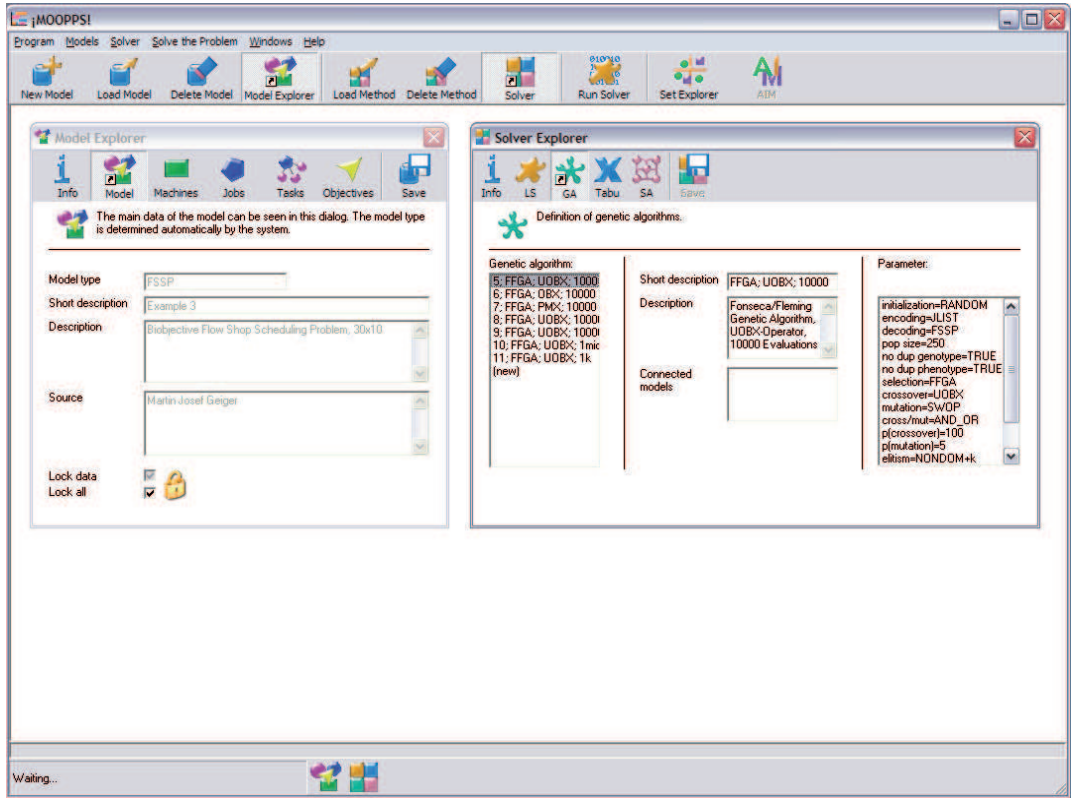}}
\caption{\label{fig:screenshot:1}Screenshot of the user
interface.}
\end{figure}

The \emph{model instance database} stores the data of the problem
instances that have to be solved. General job shop as well as flow
shop scheduling problems can be formulated. Besides newly
generated data sets, well-known test instances from literature
\cite{beasley:1996:article} have been included. Solutions are
obtained by linking model instances with methods. This allows the
reuse of specific metaheuristics for a range of problem instances
as well as the comparison of results obtained from different
heuristic approaches. A graphical \emph{user interface} as given
in Figure \ref{fig:screenshot:1} links the modules described above
into a single system.

\subsection{Optimization and decision making}
The resolution of multi objective scheduling problems is supported
by a procedure consisting of two stages. First, Pareto optimal
alternatives or an approximation $P_{a}$ of the Pareto set $P$ are
computed using the chosen metaheuristics. Second, an interactive
search in the obtained results is performed by the decision maker.

\begin{figure}[ht!]
\centerline{\includegraphics[width=12cm]{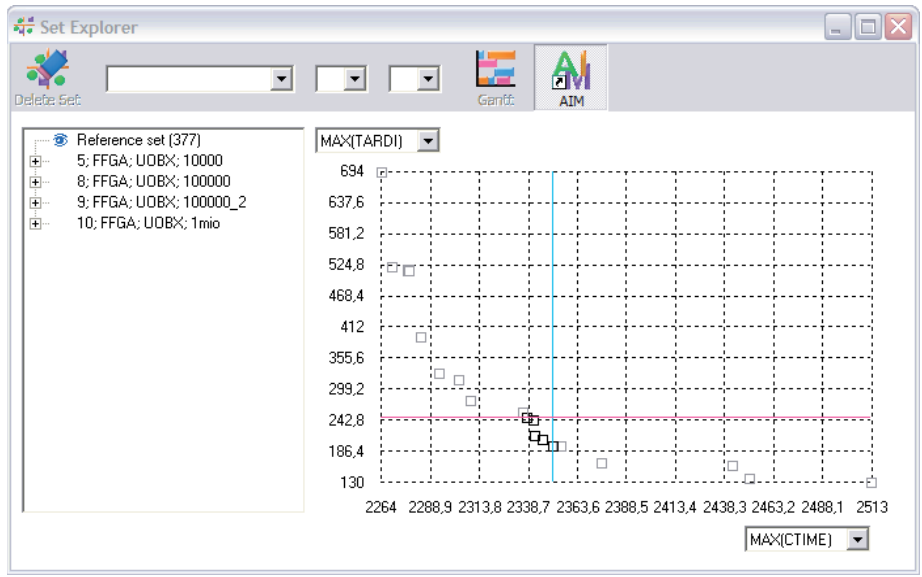}}
\caption{Dividing the approximation $P_{as}$ using aspiration
levels $a_{g_{i}}$.}\label{fig:refset}
\end{figure}

During this interactive decision making procedure, aspiration
levels $A=\{a_{g_{1}},...,a_{g_{k}}\}$ for each of the optimized
objective functions $G(x) = ( g_{1}(x),\ldots ,g_{k}(x) )$ are
chosen. As shown in Figure \ref{fig:refset}, the elements of the
approximation $P_{a}$ of the Pareto set $P$ are accordingly
divided into two subsets, the subset $P_{as}$ of the alternatives
fulfilling the aspiration levels ($g_{i}(x)\leq a_{g_{i}}\forall
i=1,...,k$) and the subset $P_{\neg as}$ of the alternatives that
do not meet the aspiration levels. It is obvious that $P_{as}\cup
P_{\neg as} = P_{a}$ and $P_{as}\cap P_{\neg as}=\emptyset$.

The initial values of the aspiration levels $a_{g_{i}}$ are set to
the worst values in $P_{a}$: $a_{g_{i}}=\max\limits_{x\in
P_{a}}(g_{i}(x))\forall i=1,...,k$ and as a consequence,
$P_{as}=P_{a}$. The decision maker is allowed to modify the values
of the aspiration levels and successively reduce the number of
elements in $P_{as}$ until $\mid P_{as} \mid =1$. The remaining
alternative in $P_{as}$ is the desired compromising solution
$x^{*}$ as the fixed aspiration levels are met by this
alternative.

\begin{figure}[ht!]
\centerline{\includegraphics[width=14cm]{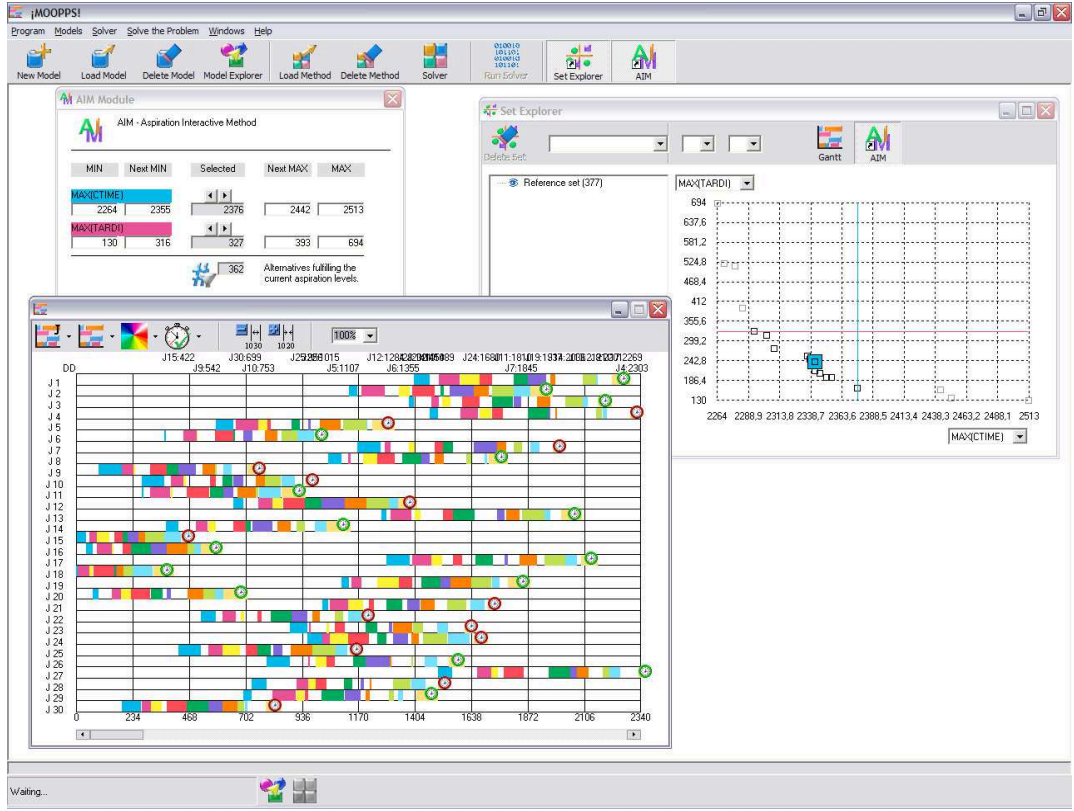}}
\caption{\label{fig:screenshot:2}Screenshot of Gantt presentation
of the solution.}
\end{figure}

As Figure \ref{fig:screenshot:2} demonstrates, the decision maker
does not only have to rely on the choice of aspiration levels but
is able to visualize the corresponding Gantt chart of a particular
schedule.

\section{Conclusions}
A decision support system for multiple objective scheduling
problems has been presented. It incorporates a set of
metaheuristics that can be adapted to specific problems instances.
As the user interface is highly visual, nonexperienced users are
able to solve scheduling problems under multiple objectives with
comparably little knowledge.

After an approximation of Pareto optimal alternatives has been
obtained, an interactive decision making module based on the
aspiration interactive method allows the identification of a most
preferred schedule. The system may also be used to compare
different approximation results of various metaheuristic
approaches in terms of their approximation quality. It is
therefore suitable for demonstrating the use, adaptation and
effectiveness of metaheuristics to complex combinatorial
optimization problems using the example of machine scheduling
under multiple objectives.

\section*{Acknowledgements}
The author would like to thank Zs\'{i}ros \'{A}kos (University of
Szeged), Pedro Caicedo, Luca Di Gaspero (University of Udine), and
Szymon Wilk (Poznan University of Technology) for providing
multilingual versions of the software.

\end{document}